\pdfoutput=1

\documentclass[11pt]{article}

\usepackage[dvipsnames]{xcolor}
\usepackage{acl}

\usepackage{times}
\usepackage{latexsym}

\usepackage[T1]{fontenc}

\usepackage[utf8]{inputenc}

\usepackage{microtype}

\usepackage{multirow}
\usepackage{booktabs}
\usepackage{graphicx}
\usepackage{subfigure}
\usepackage{pifont}
\usepackage{xspace}
\usepackage{tabularx}
\usepackage{amsmath}
\usepackage{tablefootnote}
\usepackage{algorithm}
\usepackage[noend]{algpseudocode}
\usepackage{amssymb}
\usepackage{hyperref}

\usepackage{soul}

\newcommand{\ttbf}[1]{\texttt{\textbf{#1}}}
\newcommand{\datafbn}[0]{\ttbf{FB15K-237N}\xspace}
\newcommand{\datafb}[0]{\ttbf{FB15K-237}\xspace}
\newcommand{\datawnrr}[0]{\ttbf{WN18RR}\xspace}
\newcommand{\dataicews}[0]{\ttbf{ICEWS14}\xspace}
\newcommand{\datanell}[0]{\ttbf{NELL-One}\xspace}
\newcommand{\method}[0]{\ttbf{KG-S2S}\xspace}

\title{Knowledge Is Flat: A Seq2Seq Generative Framework for Various Knowledge Graph Completion}

\author{Chen Chen$^{1,*}$, Yufei Wang$^{2,}$\thanks{~~First two authors contribute equally.}, Bing Li$^3$ \and Kwok-Yan Lam$^{1}$\thanks{~~Corresponding author} \\
Nanyang Technological University, Singapore$^1$ \\
Macquarie University, Sydney, Australia$^2$ \\
Centre for Frontier AI Research (CFAR), A*STAR, Singapore$^3$\\
\texttt{\{S190009,kwokyan.lam\}@ntu.edu.sg},\\
\texttt{yufei.wang@students.mq.edu.au} \\
\texttt{li\_bing@ihpc.a-star.edu.sg} \\
}

\begin{document}
\maketitle
\begin{abstract}
Knowledge Graph Completion (KGC) has been recently extended to multiple knowledge graph (KG) structures, initiating new research directions, e.g. static KGC, temporal KGC and few-shot KGC~\cite{survey}. Previous works often design KGC models closely coupled with specific graph structures, which inevitably results in two drawbacks: 1) structure-specific KGC models are mutually incompatible; 2) existing KGC methods are not adaptable to emerging KGs. In this paper, we propose \method, a Seq2Seq generative framework that could tackle different verbalizable graph structures by unifying the representation of KG facts into ``flat'' text, regardless of their original form. To remedy the KG structure information loss from the ``flat'' text, we further improve the input representations of entities and relations, and the inference algorithm in \method. Experiments on five benchmarks show that \method outperforms many competitive baselines, setting new state-of-the-art performance. Finally, we analyze \method's ability on the different relations and the \emph{Non-entity Generations}~\footnote{Our source code is available at~\url{https://github.com/chenchens190009/KG-S2S}}. 

\end{abstract}

\section{Introduction}
\begin{figure}
    \centering
    \includegraphics[width=\linewidth]{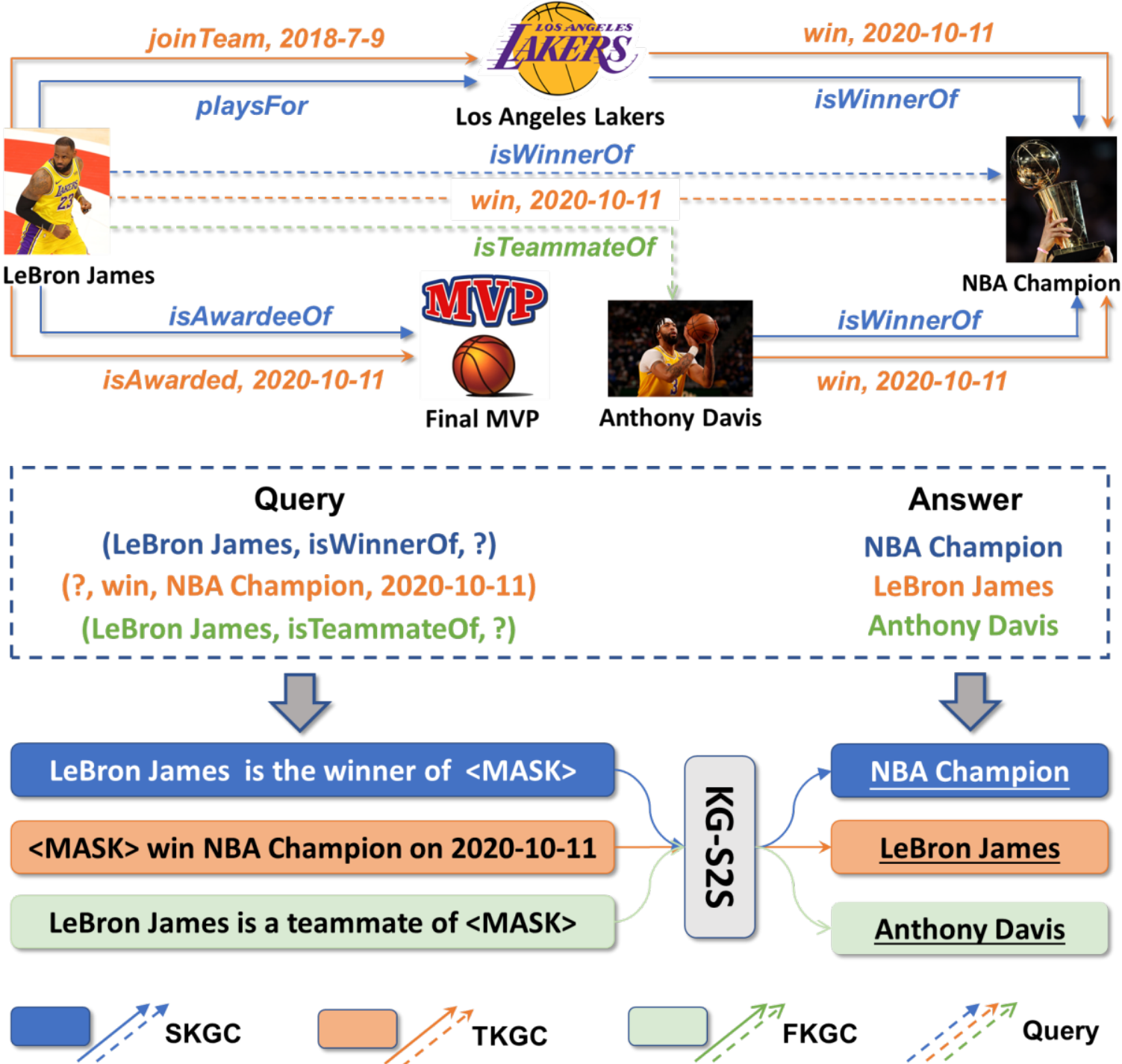}
    \caption{Running examples of Static (SKGC), Temporal (TKGC) and Few-shot (FKGC) Knowledge Graph Completion tasks. Our proposed \method is an unified Seq2Seq framework adaptable to all of these tasks.}
    \label{fig: intro}
\end{figure}

Knowledge graph completion (KGC) has been a fundamental task to discover unobserved facts from various knowledge graph (KG) structures, including static KGC (SKGC), temporal KGC (TKGC) and few-shot KGC (FKGC) ~\cite{survey}. 
As shown in Figure~\ref{fig: intro}, TKGC (in \textcolor{orange}{orange}) contains temporal facts with timestamps, while FKGC (in \textcolor{LimeGreen}{green}) predicts the facts with relations that only have limited or zero training instances. 

Typically, the solutions for KGC are graph-based, i.e., treating entities and relations as \textit{nodes} and \textit{linkages}. 
The training and inference of SKGC models rely on various \emph{transitional relations} over graph paths~\cite{ComplEx, ConvE, CompGCN}. 
TKGC and FKGC methods are further integrated with non-trivial components or learning paradigms to handle the extra temporal information or training requirements. Concretely, TKGC models~\cite{HyTE,DE-SimplE,TNTComplEx} either construct temporal-specific sub-KG or add additional temporal embeddings into existing SKGC methods. FKGC models apply the additional training scheme (e.g., meta-learning) between the frequent relations and the infrequent ones to the SKGC models~\cite{GMatching, MTransH}. 
Such a methodological discrepancy leads to a great maintenance cost and being inadaptable to emerging knowledge queries, ingestion, and presents. 
Naturally, a research question has been raised: \emph{Can we adapt the different forms of KG facts and solve these KGC tasks in a unified framework?}

Recently, Seq2Seq Pre-trained Language Models (PLM) have shown state-of-the-art performances and high technical homogeneity when dealing with different NLP tasks. Albeit having heterogeneous input and output, the Seq2Seq PLMs covert those tasks into ``text-to-text'' format, taking the text as inputs and producing another text as outputs~\cite{JMLR:v21:20-074,xie2022unifiedskg}. In addition, PLMs have embedded massive real-world knowledge from the pre-training ~\cite{petroni-etal-2019-language}, which is potentially beneficial for the KGC tasks, especially in the data-sparsity scenarios. 

Inspired by this, we propose \method, a simple yet effective Seq2Seq PLM framework adaptable to various KG structures. Given a KG query, \method directly generates the target entity text using the common PLM fine-tuning practices. Firstly, to remedy the KG structure information loss caused by the na\"ive ``text-to-text'' format, we improve \method via 1) the input representations of entities and relations using \emph{Entity Description}, \emph{Soft Prompt} and \emph{Seq2Seq Dropout}; 2) the constrained inference algorithm empowered by the \emph{Prefix Constraints}; Secondly, 
we treat all the KG elements (i.e., entity, relation and timestamp) as ``flat'' text (Figure~\ref{fig: intro}) which enables \method to \textbf{i)} handle various \emph{verbalizable} knowledge graph structures; \textbf{ii)} generate non-entity text and find novel entities for KGs. We make several improvements on the preliminary attempts of concurrent works~\cite{KGT5, GenKGC} using Seq2Seq for KGC. Our model adds special treatments to input entity/relation textual representation. This helps to better capture subtle yet key tokens and facilitate the ability to ingest other graph structures.

We conduct experiments on \datawnrr, \datafb and \datafbn for SKGC, \dataicews for TKGC and \datanell for FKGC. \method outperforms several competitive baseline models, including graph-based and PLM-based models, and sets new state-of-the-art performance on all three settings. We conduct ablation studies to show the effectiveness of the proposed components, compare \method with graph-based KGC models at the relation level and finally showcase the \emph{Non-entity Generation} from \method to present its potential in producing novel knowledge triples.

\section{Related work}
KGC has been studied in the Static, Temporal and Few-shot settings. Previous works often focus on a single setting, while \method fits all three settings without any architecture modifications.

\paragraph{Static KGC} Early SKGC models assign trainable embeddings to each entity and relations~\cite{TransE, RotatE}. A score function is proposed to evaluate the scores of triples with these embedding. These models learn structural information of a knowledge graph, regardless of the textual information of the entities and relations. Recently, ~\citet{KG-BERT, StAR, GenKGC, KGT5} proposed to encode entity and relation textual knowledge into the model by using PLMs. Instead of calculating scores from embeddings, they train PLMs to produce plausibility scores for KG text representation.   


\paragraph{Temporal KGC} Many TKGC models incorporate additional time-specific parameters upon existing KGC methods.~\citet{TTransE}, based on~\citet{TransE}, represents each timestamp with independent embeddings. ~\citet{HyTE} resembles ~\citet{TransH}, regarding timestamps as hyperplanes for entities to project. ~\citet{TNTComplEx} considers the score of each triple as canonical decomposition of order 4 tensors in complex domain. ~\citet{DE-SimplE} suggests learning dynamic embeddings for entity and relations, transforming part of the embedding with sinusoidal activation of learned frequencies. ~\citet{TKGCframework} proposes a systematic framework to improve existing temporal embedding models. 


\paragraph{Few-shot KGC} For one-shot learning on relations, ~\citet{GMatching} attempts to seek a matching metric that can be used to discover similar triples given one reference triple. ~\citet{MetaR} discovers two kinds of relation-specific meta-information: relation meta, and gradient meta. It uses meta-learning methods to transfer meta-information to low-resource relations. With the help of textual information and PLMs,~\citet{StAR} outperforms other few-shot baseline models on the zero-shot relations.


\begin{figure*}[!ht]
    \centering
    \includegraphics[width=\linewidth]{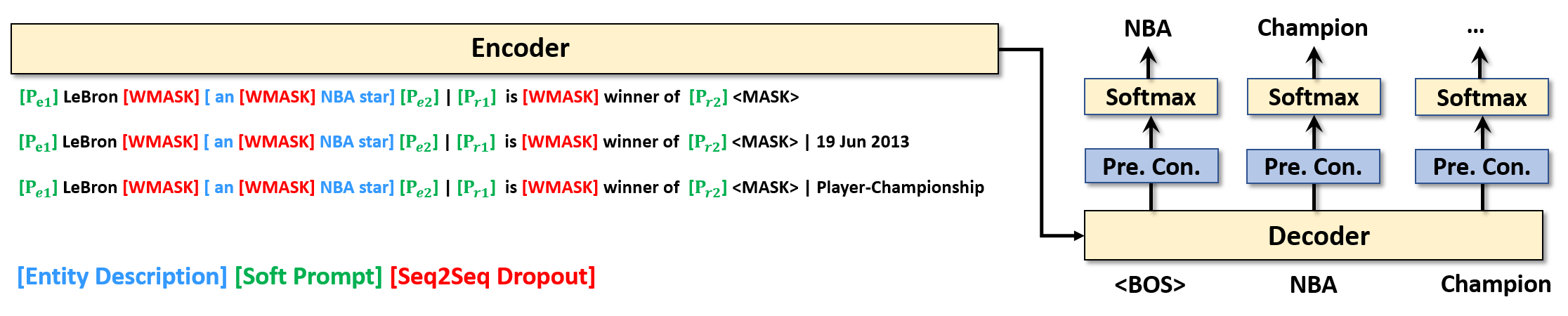}
    \caption{The overview of \method. Given the query (LeBron James, is the winner of, ?, $*$), we represent ``LeBron James'' by joining entity name and description. \emph{Soft Prompt} \textit{distinguishes} similar relation name and \textit{disentangles} relation-specific information. \emph{Seq2Seq Dropout} randomly masks input words to avoid over-fitting. In the inference, \emph{Prefix Constraints }(Pre. Con.) forces the decoding algorithms to \textit{only} generate valid entity text in the decoder.}
    \label{fig:method}
\end{figure*}

\section{Proposed Method}
This section first formulates \emph{Knowledge Graph Completion} tasks in Sec.~\ref{task}, then discusses our proposed \method method in Sec.~\ref{s2sbackground},~\ref{representation} and~\ref{inference}. Figure~\ref{fig:method} shows the overview of \method.


\subsection{Knowledge Graph Completion}
\label{task}
A \emph{Knowledge Graph} (KG) ($\mathcal{E}$, $\mathcal{R}$, $\mathcal{T}$) includes an entity set $\mathcal{E}$, a relation set $\mathcal{R}$ and a tuple list $\mathcal{T} = [(h, r, t, m)_1, \cdots, (h, r, t, m)_n]$ where $h$, $t \in \mathcal{E}$ is head and tail entity, $r \in \mathcal{R}$ is the tuple relation and $m$ is the KG meta-information. \textit{Knowledge Graph Completion} (KGC) predicts the missing entities for the queries $(?, r, t, m)$ or $(h, r, ?, m)$. 

The meta-information $m$ denotes different form of contents in different KG settings. As shown in Figure~\ref{fig:method}, $m$ is represented as null in SKGC, timestamps (e.g., ``Jun-19-2014'') in TKGC and typing (e.g., Player-Championship) in the KGs providing typing information. Using text representation, KGs with different structures can be converted into an unified format.


\subsection{A Seq2Seq Framework for KGC}
\label{s2sbackground}

A Seq2Seq Framework, including an \emph{encoder} and a \emph{decoder}, can be viewed as:
\begin{align}
P(Y|X) = \prod_{t=1}^m P(y_t|X, Y_{<t})
\end{align}
where $X$ is the input sequence to the Seq2Seq \emph{encoder}, $Y$ is the auto-regressively generated output sequence (i.e., from left to right) and $y_0$ is the special Begin-of-Sequence Symbol. To apply this Seq2Seq Framework to KGC, given query $(?, r, t, m)$ or $(h, r, ?, m)$ and corresponding ground-truth answer $gt$, we first encode $r$, $t$ and $m$ into text. We represent ``?'' with ``<mask>'' at the corresponding position to distinguish between $(?, r, t, m)$ and $(h, r, ?, m)$. We then concatenate the text together into $X$ and train \method to generate $gt$ as output sequence $Y$. The \method training is straightforward: unlike StAR~\cite{StAR} which applied composite objective over the encoder-only PLM, we follow the common practices in fine-tuning Seq2Seq PLMs (i.e., Cross-Entropy Loss), directly training \method with positive examples (negative sampling trick is unnecessary to \method). However, this architecture remains two main challenges: \emph{i)} How to \emph{effectively} represent the query in the \method encoder? \emph{ii)} How to \emph{accurately} generate entity text as the answer to the query? Sec.~\ref{representation} and Sec~\ref{inference} answer the above two questions, respectively.

\subsection{Entity \& Relation Representation}
\label{representation}
Encoding query $(?, r, t, m)$ and $(h, r, ?, m)$ into ``flat'' text allows \method to handle various KGs. However, the ``flat'' text could introduce KG structure loss. To remedy this issue, we further improve \method using the following components.


\paragraph{Entity Description}
Intuitively, one could represent an entity using its name text (e.g., Anthony Davis) in either compositional or non-compositional form~\cite{li2018adaptive}. However, as \method is initialized from the PLM weights, some specific types of entities (e.g., person, locations) may refer to multiple real-world entities in the large-scale PLM training corpus, introducing noisy ambiguity to \method. To avoid this risk, we additionally introduce entity descriptions to enrich the context information of entities. For example, the textual description about ``Lebron James'' could be ``is an American NBA star''. Previous research ~\cite{DKRL,lovelace-etal-2021-robust} have shown the utility of the descriptions when integrated with traditional graph-based KGC models. Likewise, we add entity descriptions for both queries and ground-truth answers. At the encoder side, we concatenate entity names and descriptions as the entity representation. At the decoder side, we \emph{train} \method to jointly predict entity names and entity descriptions under the cross-entropy loss. We find using entity descriptions on both sides of \method is beneficial.

\paragraph{KG Soft Prompt}
In traditional graph-based KGC models (e.g., TransE), KG entities and relations are represented with separated embeddings, while \method represents entities and relations using the shared Seq2Seq PLM parameters. As a consequence, the KG knowledge/patterns regarding similar surface relations or entities (e.g., ``film costume design by'' and ``film production design by'') could be mixed together.  To tackle this issue and inspired by the recent \emph{Soft Prompt}~\cite{lester-etal-2021-power}, which is a set of trainable embeddings directly fed into the Seq2Seq PLM input, we propose to add additional trainable prompt embeddings for specific entities and relations into $X$. The separated parameter space could potentially \emph{disentangle} the general KG and element-specific knowledge for \method. However, as entities are equipped with descriptions and \emph{Entity Soft Prompt} introduces a large number of parameters, we only apply the \emph{Soft Prompt} to relations. Specifically, as shown in Figure~\ref{fig:method}, similar to a recent BERT-based KGC model~\citep{fb15k-237n}, we insert the \emph{Relation Soft Prompt} embeddings $P_{e1}, P_{e2}, P_{r1},  P_{r2} \in \mathbb{R}^{|\mathcal{R}| \times d}$, where $d$ is the \method hidden size, before and after the textual entity and relation name.


\paragraph{Seq2Seq Dropout}
In our preliminary experiments, we find that \method often learns fast (measured by validation MRR) in the early stage of the model training. We hypothesize that, unlike other NLG tasks where different instances have little textual overlapping, the entity descriptions remain unchanged in different training queries in KGC, which could easily lead to over-fitting. We attempt to increase the original \emph{encoder dropout} for \method training, however, it has little impact on the final performance. Therefore, we impose a more strict \emph{Seq2Seq Dropout} where we randomly select and mask $p$\% of the input tokens in $X$ when calculating the encoder self-attention module and decoder cross-attention module. Note that the \emph{Relation Soft Prompt} and the ``<mask>'' token are excluded from this selection process. Compared with the original \emph{encoder dropout}, Seq2Seq dropout takes effect at both encoder and decoder sides. This introduces more diversity to the input query text, thus, better capability of preventing over-fitting.



\subsection{KGC Inference}
\label{inference}
The traditional KGC models $g$ answer a query $(?, r, t, m)$  by first finding the score $g(x, r, t, m)$ $\forall x \in \mathcal{E}$ and then ranking all entities based on $g(x, r, t, m)$. Naturally, at the inference stage, \method could compute a score for every $x \in \mathcal{E}$. However, this could be computationally expensive because $|\mathcal{E}|$ could be very large (e.g., $|\mathcal{E}|$ is 68,544 in \datanell). Instead, 
in \method, its encoder takes $X$ as input and then the \method decoder generates the text of entity predictions that are mapped into specific entity ids. These generated entities are further ranked based on their corresponding log cross-entropy loss. We assign -$\infty$ for all entities not generated in the decoding stage.

\paragraph{Decoding Methods}
Different from general text generation tasks where only one optimal output sequence is required, in KGC, given a query $(?, r, t, m)$ or $(h, r, ?, m)$, there could be multiple valid entities. To generate $K$ valid entity candidates, we deploy the standard beam search algorithm with beam width $K$ because it naturally produces different entity text in each beam with high likelihood. In contrast, random sampling often provides low-quality answers due to its randomness in decoding and the outputs of diverse beam search are distorted due to its diversity encouragement term.

\def\gT{{\mathcal{T}}}
\def\gPT{{\mathcal{PT}}}

\begin{algorithm}[t!]
\caption{Next Candidates (NC): Given Entity Prefix Trie $\gT$, Query-GT Prefix Trie Mapping $\mathcal{D}$, query $q$ and the generated tokens $Gen$; return candidate tokens.
}
\label{alg:prefix_trie}
\begin{algorithmic}[1]
\Procedure{NC}{$\gT, \mathcal{D}, q, Gen$}
    \State $\mathcal{T}_{q} \gets \mathcal{D}.get(q)$
    \State $cand \gets \gT.next(pre=Gen)$
    \State $rm \gets \mathcal{T}_{q}.next(pre=Gen)$

    \State $cand \gets cand.remove(rm)$
    \State \textbf{return} $cand$
\EndProcedure
\end{algorithmic}
\end{algorithm}
\paragraph{Prefix Constraints}
The flexible auto-regressive generation may produce entities that do not exist in $\mathcal{E}$, which could reduce the number of valid entity candidates in the decoding. To avoid this scenario, we propose \emph{Prefix Constraints} to control the \method decoder to generate valid tokens given prefix sequences $p$. For example, given $\mathcal{E}$ = \{``Grammy Award for Best Rock Song'', ``Grammy Award for Best Music Video''\} and $p$ = [Grammy, Award, for, Best], the \emph{Prefix Constraints} only allow ``Rock'' and ``Music'' to be generated in the next step. To enable effective decoding, we propose to use Trie~\cite{cormen2009introduction} to extract appropriate next tokens. As suggested in Algorithm~\ref{alg:prefix_trie}, given the generated prefix, we first extract all possible tokens using the Entity Prefix Trie $\mathcal{T}$ and then remove the entities that are the ground-truth to the query $q$ in the training data using the Query-GT Prefix Trie $\mathcal{T}_{q}$.




\begin{table*}[!htbp]
	\centering
	\begin{small}
	\resizebox{\textwidth}{!}{
	\begin{tabular}{lcccccccccccccccc}
		\toprule
		& \multicolumn{4}{c}{\textbf{\datawnrr{}}} &
		\multicolumn{4}{c}{\textbf{\datafb{}}} &
		\multicolumn{4}{c}{\textbf{\datafbn{}}} &  \\ 
		\cmidrule(r){2-5}  \cmidrule(r){6-9} \cmidrule(r){10-13} 
&MRR &H@1 &H@3 &H@10 &MRR &H@1 &H@3 &H@10 &MRR &H@1 &H@3 &H@10 \\
		\midrule
\textbf{\emph{\footnotesize{Graph-Based Methods}}} \\
TransE~\cite{TransE} &.243 &.043 &.441 &.532 &.279 &.198 &.376 &.441 &.255 &.152 &.301 &.459\\
DistMult~\cite{DistMult} &.444 &.412 &.470 &.504 &.281 &.199 &.301 &.446 &.209 &.143 &.234 &.330 \\
ComplEx~\cite{ComplEx} &.449 &.409 &.469 &.530 &.278 &.194 &.297 &.450 &.249 &.180 &.276 &.380\\
ConvE~\cite{ConvE} &.456 &.419 &.470 &.531 &.312 &.225 &.341 &.497 &.273 &.192 &.305 &.429\\
RotatE~\cite{RotatE} &.476 &.428 &.492 &.571 &.338 &.241 &.375 &.533 &.279 &.177 &.320 &.481\\
CompGCN~\cite{CompGCN} &.479 &.443 &.494 &.546 &.355 &.264 &.390 &.535  &.316 &.231 &.349 &.480\\
\midrule
\textbf{\emph{\footnotesize{PLM-Based Methods}}} \\ 
KG-BERT~\cite{KG-BERT} &.216 &.041 &.302 &.524 &- &- &- &.420 &.203 &.139 &.201 &.403\\
MTL-KGC~\cite{MTL-KGC} &.331 &.203 &.383 &.597 &.267 &.172 &.298 &.458 &.241 &.160 &.284 &.430\\
StAR~\cite{StAR} &.401 &.243 &\underline{.491} &\textbf{.709} &\underline{.296}  &.205 &.322 &\underline{.482} &- &- &- &- \\
PKGC~\cite{fb15k-237n} &- &- &- &- &- &- &- &- &\underline{.307} &\underline{.232} &\underline{.328} &\underline{.471} \\
GenKGC~\cite{GenKGC} &- &.287 &.403 &.535 &- &.192 &\underline{.355} &.439 &- &- &- &- \\
KGT5~\cite{KGT5} &\underline{.508} &\underline{.487} &- &.544 &.276 &\underline{.210} &- &.414 &- &- &- &- \\
\midrule
\method (Ours)	&\textbf{.574} &\textbf{.531} &\textbf{.595} &\underline{.661} &\textbf{.336} &\textbf{.257} &\textbf{.373} &\textbf{.498} &\textbf{.353} &\textbf{.282} &\textbf{.385} &\textbf{.495} \\
\bottomrule
\addlinespace
\end{tabular}
}
\caption{Results of static KGC. \datawnrr{} and \datafb{} results are taken from ~\citet{StAR}. \datafbn{} results are taken from ~\cite{fb15k-237n}. The uncovered results of graph-based methods are obtained through hyperparameter tuning with LibKGE~\cite{LibKGE} and PLM-based methods through official implementations. The best PLM-based method results are in bold and the second best results are in underline.}
\label{tab:static KG} 
\end{small}
\end{table*}

\section{Experiment}
In this section, we evaluate \method against competitive baselines in the following KGC datasets: \datawnrr~\cite{ConvE} (SKGC), \datafb~\cite{FB15k237} (SKGC) \datafbn~\cite{fb15k-237n} (SKGC) and \dataicews~\cite{TA-TransE} (TKGC) and \datanell~\cite{GMatching} (FKGC).

\subsection{Experimental Settings}
\paragraph{Dataset}
\datawnrr and \datafb are improved version of WN18 and FB15k~\cite{TransE} respectively, where all inverse relations are removed to avoid data leakage. \datafbn further removes \datafb 's concatenated relations caused by Freebase mediator nodes ~\cite{CVT} to avoid Cartesian production relation issue. \dataicews refers to 2014 political facts from the Integrated Crisis Early Warning System database~\cite{DVN/28075_2015}. \datanell is a few-shot KGC dataset derived from NELL~\cite{10.5555/2898607.2898816}. Following ~\citet{StAR}, we reformat \datanell so that the dev/test relations never appear in the train set. More details can be found in Appendix \ref{appendix:dataset}.

\paragraph{Implementation details}
We initialize \method using the T5-base model~\cite{JMLR:v21:20-074}, 
and optimize \method with Adam~\cite{DBLP:journals/corr/KingmaB14}. We use T5 default settings in our experiments for all benchmarks and follow the \textit{filtered setting} proposed in~\citet{TransE} to evaluate our model. More implementation details and optimal hyperparameters can be found in Appendix~\ref{appendix:implementation}.

\paragraph{Evaluation Protocol} We remove the duplicated entities from the output. In the  non-constrained decoding method, we further remove non-entity generations. The performance of our model is reported on the standard KGC metrics: Mean Reciprocal Rank (MRR), and Hits@1,3,10 (Hits@1,5,10 in \datanell to follow previous works). For each test triple $(h, r, t, m)$, we rank all entities for the query $(h, r, ?, m)$ and $(?, r, t, m)$. We then aggregate the ranking for ground-truth entity and report the mean reciprocal rank (MRR) and the proportion of ground-truth entities ranked in the top $n$ (H@$n$). To handle the equal score scenarios, we use the RANDOM mode proposed in ~\citet{reevaluation} to determine the rank of entities. Model is selected by MRR value on valid set.



\subsection{Experimental results}
\label{sec: experimental result}
\paragraph{Static KGC}
We compare our results with various graph-based and PLM-based methods on the SKGC settings. Experimental results are summarized in Table \ref{tab:static KG}. On \datawnrr and \datafb, \method achieves state-of-the-art or competitive performance. In the comparison of PLM-based methods, \method outperforms previous work by a substantial margin. Specifically, we see 13\% (from 0.508 to 0.574) relative MRR improvement on \datawnrr, and 16\% (from 0.296 to 0.336) on \datafb. Compared with graph-based methods, \method consistently obtains performance gain on \datawnrr, though maintaining a modest result on \datafb.  

\begin{table}[!htbp]
	\centering
	\resizebox{0.8\linewidth }{!}{
	\begin{tabular}{llcccc}
\toprule
 &relations &MRR &H@1 &H@3 &H@10 \\
\midrule

RotatE &CPR &.337 &.232 &.374 &.552 \\
&non-CPR &.340 &.254 &.376 &.504 \\
&all &.338 &.241 &.375 &.533 \\
\midrule

\method &CPR &.318 &.234 &.355 &.493 \\
&non-CPR &.363 &.292 &.398 &.504 \\
&all &.336 &.257 &.373 &.498 \\
\bottomrule
\end{tabular}
}
\caption{Evaluation of cartesian product relations (CPRs) and non-cartesian product relations (non-CPRs) on \datafb}
\label{tab:cartesian product relations}
\end{table}
According to ~\citet{CVT, fb15k-237n}, \datafb contains many over-simplified unrealistic cartesian product relations (CPR), which improperly improves the model accuracy. For instance, the multiary fact ``average low temperature in Tokyo is 34 degrees Fahrenheit in January'' has been decomposed into multiple CPR facts (Tokyo, climate./month, January) and (Tokyo, climate./average\_min\_temp, 34), which are obviously unrealistic and semantically meaningless. We note that RotatE achieves higher overall performance than \method on \datafb. 

However, after breaking down the performance on CPRs and non-CPRs in Table~\ref{tab:cartesian product relations}, we surprisingly find that our proposed \method has distinct advantages on non-CPR (MRR 0.363 vs. 0.338). That is, leading performance of RotatE is due to the facts with CPR, while \method has demonstrated its advantages in realistic relations (i.e., non-CPRs). Methodologically, RotatE is a typical graph-based model, while KG-S2S regards KGs as plain text with structure-aware components. Graph-based models are good at predicting simple structure yet inferior in absorbing KGs text. This explains why RotatE performs better on FB15k-237 dataset which is rich in cartesian product relations (CPRs, simple synthesized yet less textually meaningful relations), while worse on non-CPR datasets like FB15K-237N. This further motivates us to compare \method with other KGC methods on \datafbn which only has facts with non-CPR. As shown in Table \ref{tab:static KG}, \method obtains the best results compared with graph-based and PLM-based baselines at all metrics. In particular, \method achieves an absolute Hit@1 increase of 5.0\% over second best method \citet{StAR}.


The overall SKGC results confirms that, by taking advantage of entity and relation textual representation, \method is capable of capturing more accurate semantics of KG facts, and employ them for inference.

\paragraph{Temporal KGC}
\begin{table}[!htbp]
	\centering
	\resizebox{\linewidth}{!}{
	\begin{tabular}{lcccc}
		\toprule
        &MRR &H@1 &H@3 &H@10  \\
		\midrule
\textbf{\emph{\footnotesize{Graph-Based Methods}}} \\ [-0.3ex]
TTransE~\cite{TTransE} &.255 &.074 &- &.601 \\
HyTE~\cite{HyTE} &.297 &.108 &.416 &.655 \\
ATiSE~\cite{ATiSE} &.550 &.436 &\underline{.629} &.\underline{750} \\
DE-SimplE~\cite{DE-SimplE} &.526 &.418 &.592 &.725 \\
Tero~\cite{Tero} & \underline{.562} &.468 &.621 &.732 \\
TComplEx~\cite{TNTComplEx} &.560 &\underline{.470} &.610 &.730 \\
TNTComplEx~\cite{TNTComplEx} &.560 &.460 &.610 &.740 \\

T+TransE~\cite{TKGCframework} &.553 &.437 &.627 &\textbf{.765} \\
T+SimplE~\cite{TKGCframework} &.539 &.439 &.594 &.730 \\
\midrule
\specialrule{.4pt}{0pt}{0pt}
\textbf{\emph{\footnotesize{PLM-Based Methods}}} \\ [-0.3ex]
\method(Ours) &\textbf{.595} &\textbf{.516} &\textbf{.642} &.737  \\
\bottomrule
\addlinespace
\end{tabular}
}
\caption{Results of temporal KGC on \dataicews. All the results are from original papers.}
\label{tab:temporal KG} 
\end{table}
To evaluate \method's ability of handling additional meta-information in KG, we conduct the experiment on the TKGC benchmark \dataicews. The results are shown in Table \ref{tab:temporal KG}. Our proposed \method obtains a new state-of-the-art result on MRR and Hit@1,3 while achieving comparative performance on Hit@10. This result confirms that \method can learn additional temporal meta-information from pure textual form. We observe that our result on Hit@10 is lower than several existing methods. This could be explained by the low quality of entities in \dataicews, which only includes the ``sector'' and ``country'' of the entities. These entity descriptions are much less informative than the ones in the SKGC benchmark. We believe that the performance of \method could be further improved when more informative entity descriptions are available.


\paragraph{Few-shot KGC}
\begin{table}[!htbp]
	\centering
	\resizebox{\linewidth}{!}{
	\begin{tabular}{lccccc}
		\toprule
        &N-Shot &MRR &H@1 &H@5 &H@10  \\
		\midrule
\textbf{\emph{\footnotesize{Graph-Based Methods}}} \\ [-0.3ex]
${\mathrm{GMatching}_\mathrm{ComplEx}}^\clubsuit$ &Five &.20 &.14 &.26 &.31 \\
${\mathrm{MetaR}}^\heartsuit$ &Five &.26 &.17 &.35 &.44\\
${\mathrm{GMatching}_\mathrm{TransE}}^\clubsuit$ &One &.17 &.12 &.21 &.26\\
${\mathrm{GMatching}_\mathrm{DistMult}}^\clubsuit$ &One &.17 &.11 &.22 &.30\\
${\mathrm{GMatching}_\mathrm{ComplEx}}^\clubsuit$ &One &.19 &.12 &.26 &.31\\
${\mathrm{MetaR}}^\heartsuit$ &One &.25 &.17 &.34 &.40\\
${\mathrm{MTransH}}^{\triangle}$ &One &\textbf{.31} &\underline{.21} &\textbf{.41} &\underline{.48} \\
\specialrule{.4pt}{0pt}{0pt}
\textbf{\emph{\footnotesize{PLM-Based Methods}}} \\ [-0.3ex]
${\mathrm{StAR}}^{\spadesuit}$ &Zero &.26 &.17 &.35 &.45\\
\method (Ours) &Zero &\textbf{.31} &\textbf{.22} &\textbf{.41} &\textbf{.49} \\
\bottomrule
\addlinespace
\end{tabular}
}
\caption{Results of few-shot KGC on NELL-One. $\triangle$~\cite{MTransH}. $\clubsuit$~\cite{GMatching}, $\heartsuit$~\cite{MetaR} and $\spadesuit$~\cite{StAR}.}
\label{tab:few-shot KG} 
\end{table}
Finally, we verify \method's ability in few-shot learning in the \datanell benchmark, as shown in Table~\ref{tab:few-shot KG}. Following~\citet{StAR}, we conduct the evaluation under zero-shot setting (i.e, evaluation relations never appear in the training set). Surprisingly, \method is able to achieve superior performance than all the variations of previous graph-based models, which transfer knowledge from the training data to the evaluation relations (i.e., one-shot and five-shot meta learning). In addition, compared with the PLM-based StAR model, \method also obtains higher performance with considerable margins in terms of all the metrics. In particular, Hit@1 performance is boosted from 0.17 to 0.22, around 29\% relative improvement. This remarkable performance gain could own to the following aspects: 1) the prior knowledge contained in PLM; 2) \method's capability to transfer the knowledge from the training relations to the unseen ones.


\begin{table}[!htbp]
\centering
\resizebox{\linewidth}{!}{
\begin{tabular}{wl{1cm}cccccc|cc}
\toprule
&\multirow{2}{*}{PW} & \multicolumn{2}{c}{Description}   &\multicolumn{2}{c}{Soft Prompt} &\multirow{2}{*}{S2S. Drop} &\multirow{2}{*}{MRR} &\multirow{2}{*}{H@10}  \\
\cmidrule(lr){3-4} \cmidrule(lr){5-6}
& & SRC & TGT  & REL & ENT & & &\\
\midrule
Baseline &\ding{51} &- &- &- &- &- &.280 &.416 \\

&\ding{51} &\ding{51} &-   &- &- &- &.326 &.453 \\
&\ding{51} &\ding{51} &\ding{51}  &- &- &- &.350 &.478 \\
&\ding{51} &\ding{51} &\ding{51}  &\ding{51} &- &- &.350 &.486 \\
&\ding{51} &\ding{51} &\ding{51}  &- &\ding{51} &- &.338 &.468 \\
\midrule
KGT5 &- &\ding{51} &\ding{51}  &- &- &- &.226 &.335 \\
 &- &\ding{51} &\ding{51}  &\ding{51} &- &\ding{51} &.233 &.341 \\
\midrule
\method &\ding{51} &\ding{51} &\ding{51}  &\ding{51} &- &\ding{51} &\textbf{.353} &\textbf{.495} \\
\bottomrule
\end{tabular}
}
\caption{Ablation for the \method Input Components on \datafbn. PW denotes pretrained weight. SRC and TGT denote source and target description. REL and ENT denote relation-specific and entity-specific soft prompts. S2S.Drop denotes Seq2Seq dropout.}
\label{tab:ablation study} 
\end{table}

\subsection{Ablation Study}
\begin{figure}[!ht]
\centering
\includegraphics[width=\linewidth]{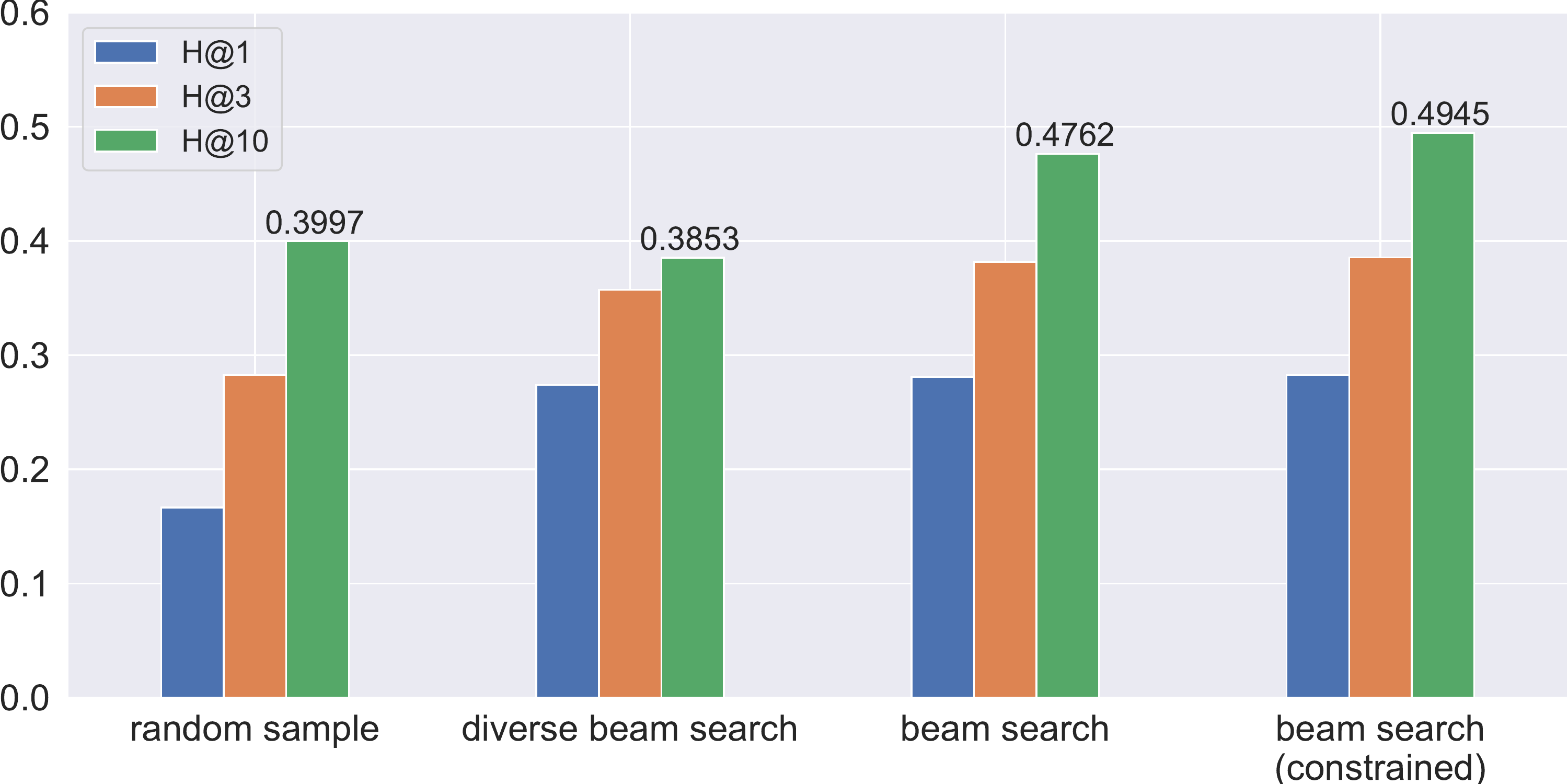}
\caption{Comparison for decoding methods on \datafbn.}
\label{fig:decoding method}
\end{figure}

In this section, we conduct ablation studies to show the contributions of each of our proposed components. Table~\ref{tab:ablation study} shows the impact of input components and Figure~\ref{fig:decoding method} shows the ablation study of decoding components in \method .




\paragraph{Source and target description} The descriptions of entities enrich their context information and resolve the ambiguity issue. As shown in Table~\ref{tab:ablation study}, adding source and target description can separately improve MRR (i.e., from 0.280 to 0.350) and Hit@10 (i.e., from 0.416 to 0.478). This suggests that to achieve optimal performance, it is important to inject entity descriptions at \method's encoder and decoder, \emph{simultaneously}.

\paragraph{Soft Prompt}
Soft prompt allows \method to recognize entities and relations as atomic concepts. Adding relation Soft Prompt successfully boosts Hit@10 from 0.478 to 0.486. However, the Entity Soft Prompt has a negative effect for \method, degrading MRR and Hit@10 by 0.12 (from 0.350 to 0.338) and 0.1 (0.478 to 0.468), respectively. We argue this phenomenon occurs for at least two reasons: 1) The entity descriptions have already enriched the entity context information and consequently made entities distinguishable; 2) Entity Soft Prompt introduces massive amounts of embeddings, which may weaken \method's ability to learn from natural language. 

\paragraph{Seq2Seq Dropout} Seq2Seq dropout applies a random masking mechanism on the encoder input mask. It is observed that Seq2Seq dropout is able to deliver consistent improvement on both MRR and Hit@10. This result practically justifies the effectiveness of this implementation. We believe the advance is derived from the diversified input data generated by Seq2Seq dropout, which helps \method to avoid potential over-fitting risk.  

\paragraph{Campared with KGT5} KGT5 is trained on a random initialized Seq2Seq structure to fully adapt KG training data. However, learnt from large pretraining corpus, pretrained weights contains rich linguistic knowledge and simply dropping them may weaken the model's ability of ingesting nature language. The large performance gap between KGT5 and \method indicates pretrained weights is critical for KGC models with Seq2Seq backbones . 


\paragraph{Impact of Decoding Methods}
In Figure~\ref{fig:decoding method}, we investigate the impact of decoding methods, including random sampling, diverse beam search, beam search and \emph{Prefix Constraints} beam search (described in Sec.~\ref{inference}) in \method. The performances of random sampling and diverse beam search are much worse than the standard beam search algorithm. This is mostly because random selection in sampling and diversity encouragement terms in diverse beam search negatively affect the quality of generated entity text. Whilst standard beam search always keeps and derives the candidates with the highest beam score from \method. We find that applying our \emph{Prefix Constraints} to the beam search algorithm further improves the \method performance (i.e., 0.02 Hit@10 improvement). \emph{Prefix Constraints} control \method to only generate valid entity text with little computation overhead. 

\subsection{Discussion}
\label{discussion}

\paragraph{Comparison with previous SOTA PLM-based methods}


From Table \ref{tab:static KG} and Table \ref{tab:few-shot KG}, \method outperforms previous SOTA encoder-only StAR methods on MRR and Hit@1,3. We argue two advantages contribute to this result: 1) Pretraining / finetuning consistency. StAR employs composite training objectives at the entity level, while its backbones (i.e. BERT~\cite{BERT} and RoBERTa~\cite{RoBERTa}) are trained with token-level cross-entropy loss. This mismatch may weaken the representation ability of PLM. In contrast, \method follows the common PLM fine-tuning practices, allowing better knowledge transfer from PLM. 2) Information interaction.  StAR uses a two-branch Siamese architecture~\cite{1467314} to encode the query text and answer text as two separated vectors, and calculates their dot-production as the score for ranking. Instead of compressing the them separately, \method interacts query and answer in the cross-attention module of \method decoder, auto-regressively. With more textual exposure and interaction, \method decoder trends to predict more accurate entities.


\begin{figure}[!ht]
\centering
\includegraphics[width=\linewidth]{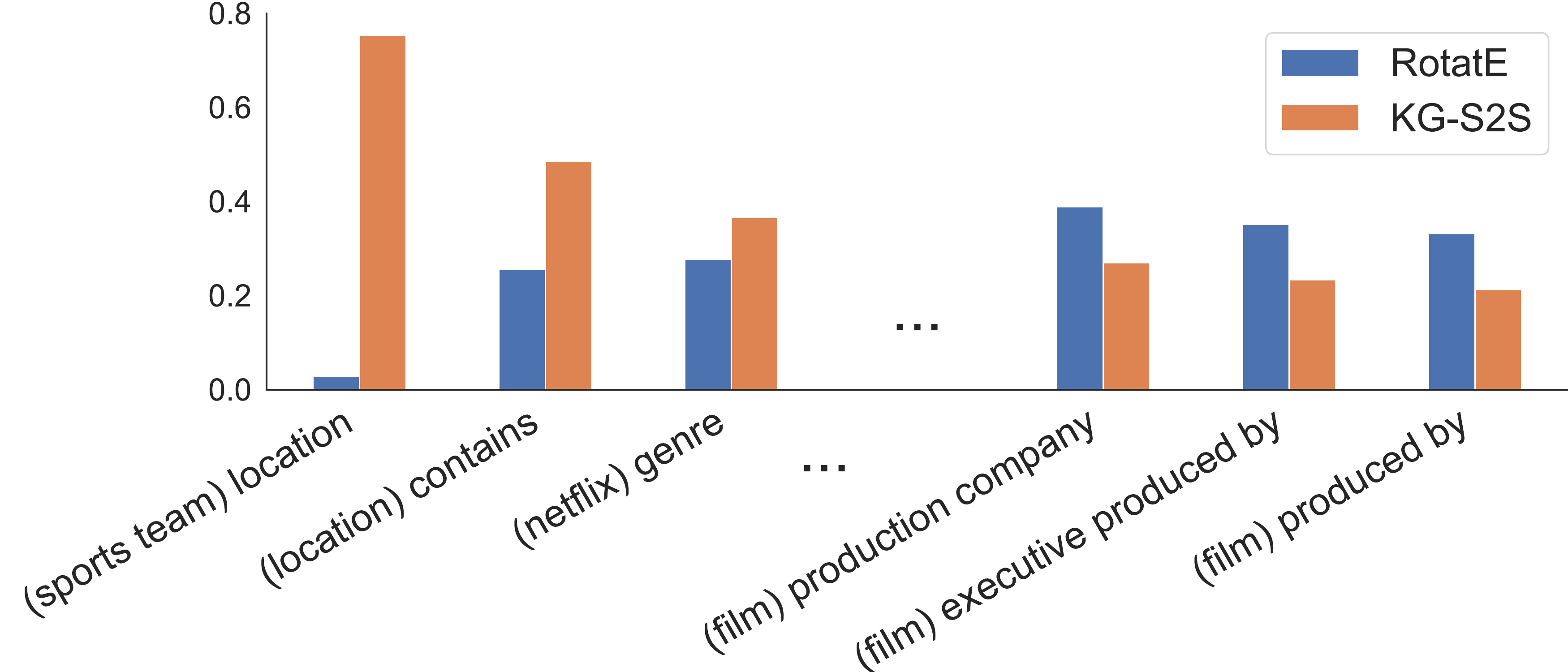}
\caption{The top-3 (left) / bottom-3 (right) relations regarding the MRR difference between \method and RotatE on the \datafbn benchmark. To maintain stable results, we select the relations with at least 0.5\% facts, and sort them by the MRR difference.}
\label{fig:relation_case_study}
\end{figure}

\paragraph{Relations Analysis} Figure~\ref{fig:relation_case_study} shows the top-3 and bottom-3 relations regarding the MRR difference between \method and RotatE. The top-3 relations are \textit{(sports team) location}, \textit{(location) contains} and \textit{(Netflix) genre}, which refer to the real-world knowledge. The possible reason is such knowledge has already been obtained from the pre-training corpus by the Seq2Seq PLM. In contrast, the bottom-3 relations are \textit{(film) production company}, \textit{(film) executive produced by} and \textit{(film) produced by}, which, surprisingly, are all relevant to the film industry. This could be because these relations are all linked to the person and company names that may have multiple references (i.e., different people and companies could share the same names) in the PLM pre-training corpus. In addition, we find that some of the relations are semantically overlapping. For example, the \datafbn includes both relation \textit{(film) executive produced by} and \textit{(film) produced by}. After being trained with the fact \textit{(Hulk, (film) executive produced by, Stan Lee)}, \method generates \textit{Stan Lee} as the top-1 candidate for the query \textit{(Hulk, (film) produced by, ?)}. However, the ground-truth entity set doesn't include \textit{Stan Lee}. This scenario has no effect on the traditional graph-based KGC models because they do not access the text at all. Similar cases also occur between \textit{(film) written by} and \textit{(film) story by}, \textit{(people) profession} and \textit{(people) specialization of}. This issue is caused by the fact that previous KGC benchmarks i) are not fully verified by experts; ii) are based on the closed-world assumption (CWA)~\cite{Keet2013}. We leave KGC benchmarks improvement as future work.




\begin{table}[!htbp]
\centering
\resizebox{\linewidth}{!}{
\begin{tabular}{ccc}
\toprule
 Queries & Prediction & GT \\
\midrule
(RoboCop, (film) genre, ? ) &Superhero film &Thriller \\
(Amber Riley, profession, ?) &Vocalist  &Actor-GB \\
(? (location) contains, Israel)  &Greater Middle East &Eurasia \\
(?, ethnicity,  M. Night Shyamalan) &Malayalam people &Indian American \\

\bottomrule
\end{tabular}
}
\caption{Case study for Non-entity generations. GT stands for ground-truth answer.}
\label{tab:non-entity_generation} 
\end{table}
\paragraph{Non-entity Generation} Without \emph{Prefix Constraints} module, \method can generate non-entity text. As shown in Table~\ref{tab:non-entity_generation}, some of the non-entity generations are also meaningful answers to the query. In the first example, Superhero film and Thriller are both semantically correct answer. In the second one, Amber Riley is actually considered as an actor and a vocalist by the public. In addition, the third and fourth examples show that \method can derive more fine-grained answers. For example, \textit{Israel} is specifically located in the \textit{Greater Middle East} and \textit{M. Night Shyamalan} is an \textit{Indian American}, born in a Malayalam-speaking Indian city. These newly generated 
entities could be potentially applied to improve the KGC model performance via a data augmentation procedure~\cite{wang2022promda}.  The expert knowledge to determine the plausibility of non-entity generations is given by the corresponding entries from Wikipedia, e.g. the Wikipedia profile for Amber Riley ~\footnote{\url{https://en.wikipedia.org/wiki/Amber_Riley}}. 

\begin{figure}[!ht]
\centering
\includegraphics[width=\linewidth]{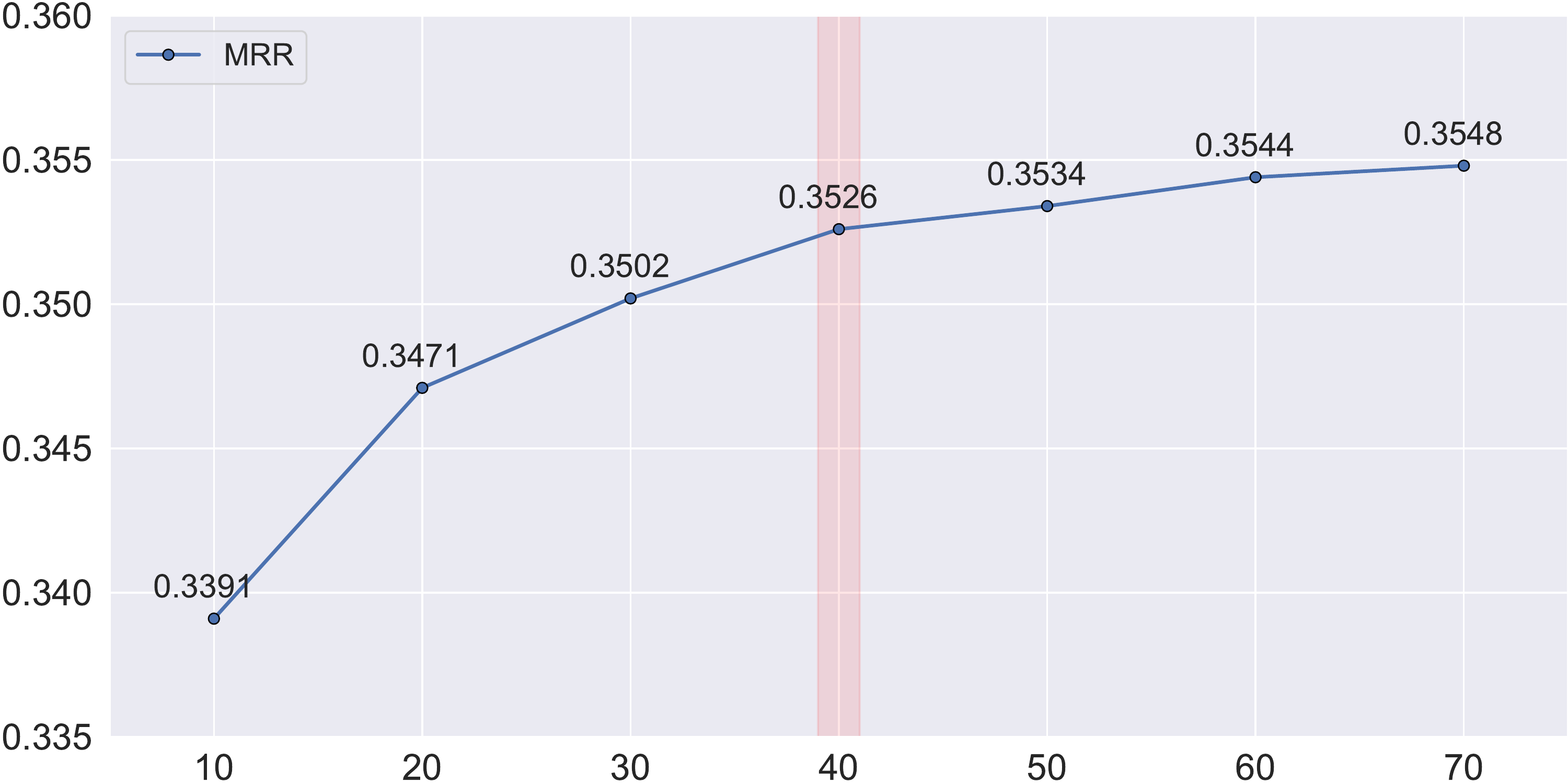}
\caption{The effect of beam width to KS-S2S}
\label{fig:beam width}
\end{figure}

\paragraph{The Effect of Beam Width}
Beam width determines the number of 
generations for each query, thus it has potentially significant impact on the \method performance. In Figure~\ref{fig:beam width}, we study how beam width affects the final performance by evaluating \method under different beam width. In general, \method achieves higher MRR as the beam width increases, whilst the performance gain becomes flat after 40 beams (red bar). As inference time goes linearly with beam width, we choose beam size 40 in \method to trade-off between model performance and inference cost.

\paragraph{Parameter Size} 
Table \ref{tab:parameters} compares the model performance and parameter size between \method and StAR. Compared with StAR (354M trainable parameters), \method is based on a smaller T5-base backbone (220M trainable parameters, 1.6x less), while it achieves better performance with a relatively large margin (MRR 0.274 vs. 0.353). We further run \method using T5-small backbone (60M trainable parameters, 5.9x less). This variant of \method obtain slightly lower result (0.351 on MRR), but still outperforms StAR with substantial margin. The results suggest \textbf{i)} \method is not sensitive to the size of PLM; \textbf{ii)} \method is parameter-efficient.

\begin{table}[!htbp]
	\centering
	\resizebox{0.8\linewidth }{!}{
	\begin{tabular}{lccc}
\toprule
Model & Size &MRR &H@10  \\
\midrule
StAR &354M &.274 &.455 \\
\method(small) &60M &.351 &.485 \\
\method(base) &220M &.353 &.495 \\
\bottomrule
\end{tabular}
}
\caption{Comparison of model performance and parameter size between \method and StAR on \datafbn.}
\label{tab:parameters}
\end{table}

\paragraph{Parameter Growth} 
Since \method represents KG elements (e.g. entities, relations and timestamps) as simple textual sequences, all KG tuples share the same  vocabulary and language model parameters. As the training KG grows, the parameters of \method only increase due to the relations \emph{Soft Prompt} with the complexity of $O(|\mathcal{R}| \cdot d)$ where $d$ is the hidden size of Seq2Seq Pre-trained language model. On the contrary, traditional graph-based models represent entities, relations and other meta-information with distinct embeddings and the parameter growth of these models is $O((|\mathcal{E}| + |\mathcal{R}|)d)$. As $|\mathcal{R}| \ll |\mathcal{E}|$, the growth could be negligible and the parameter size of \method remains nearly constant given KGs with any size.

\section{Conclusion and Future Work}
In this paper, we present \method for various knowledge graph completion tasks. By converting different kinds of KG structures into ``text-to-text'' format, \method can directly produce the text of target predicted entities. Experimental results demonstrate that \method outperforms competitive baseline models in various KGC settings. 
In the future, we would explore extending \method to other Seq2Seq PLMs, such as BART~\cite{BART} and MASS~\cite{MASS}. In addition, it is interesting to combine \method with other knowledge-intensive NLP tasks, such as conversation recommendation~\cite{NEURIPS2018_800de15c} and commonsense generation~\cite{wang-etal-2021-mention} in the Seq2Seq framework, and see if the KG knowledge could benefit these downstream tasks. 

\section*{Acknowledgement}
We thank the anonymous reviewers for their insightful suggestions to improve this paper. This work was supported by the National Research Foundation, Singapore under its Strategic Capability Research Centres Funding Initiative. Yufei Wang receives a MQ Research Excellence Scholarship and a CSIRO’s DATA61 Top-up Scholarship.

\bibliography{anthology,custom}

\begin{thebibliography}{51}
\expandafter\ifx\csname natexlab\endcsname\relax\def\natexlab#1{#1}\fi

\bibitem[{Akrami et~al.(2020)Akrami, Saeef, Zhang, Hu, and Li}]{CVT}
Farahnaz Akrami, Mohammed~Samiul Saeef, Qingheng Zhang, Wei Hu, and Chengkai
  Li. 2020.
\newblock \href {https://doi.org/10.1145/3318464.3380599} {Realistic
  re-evaluation of knowledge graph completion methods: An experimental study}.
\newblock In \emph{Proceedings of the 2020 International Conference on
  Management of Data, {SIGMOD} Conference 2020, online conference [Portland,
  OR, USA], June 14-19, 2020}, pages 1995--2010. {ACM}.

\bibitem[{Bordes et~al.(2013)Bordes, Usunier, Garc{\'{\i}}a{-}Dur{\'{a}}n,
  Weston, and Yakhnenko}]{TransE}
Antoine Bordes, Nicolas Usunier, Alberto Garc{\'{\i}}a{-}Dur{\'{a}}n, Jason
  Weston, and Oksana Yakhnenko. 2013.
\newblock \href
  {https://proceedings.neurips.cc/paper/2013/hash/1cecc7a77928ca8133fa24680a88d2f9-Abstract.html}
  {Translating embeddings for modeling multi-relational data}.
\newblock In \emph{Advances in Neural Information Processing Systems 26: 27th
  Annual Conference on Neural Information Processing Systems 2013. Proceedings
  of a meeting held December 5-8, 2013, Lake Tahoe, Nevada, United States},
  pages 2787--2795.

\bibitem[{Boschee et~al.(2015)Boschee, Lautenschlager, O'Brien, Shellman,
  Starz, and Ward}]{DVN/28075_2015}
Elizabeth Boschee, Jennifer Lautenschlager, Sean O'Brien, Steve Shellman, James
  Starz, and Michael Ward. 2015.
\newblock \href {https://doi.org/10.7910/DVN/28075} {{ICEWS Coded Event Data}}.
\newblock \emph{Harvard Dataverse}.

\bibitem[{Broscheit et~al.(2020)Broscheit, Ruffinelli, Kochsiek, Betz, and
  Gemulla}]{LibKGE}
Samuel Broscheit, Daniel Ruffinelli, Adrian Kochsiek, Patrick Betz, and Rainer
  Gemulla. 2020.
\newblock \href {https://doi.org/10.18653/v1/2020.emnlp-demos.22} {Libkge - {A}
  knowledge graph embedding library for reproducible research}.
\newblock In \emph{Proceedings of the 2020 Conference on Empirical Methods in
  Natural Language Processing: System Demonstrations, {EMNLP} 2020 - Demos,
  Online, November 16-20, 2020}, pages 165--174. Association for Computational
  Linguistics.

\bibitem[{Carlson et~al.(2010)Carlson, Betteridge, Kisiel, Settles, Hruschka,
  and Mitchell}]{10.5555/2898607.2898816}
Andrew Carlson, Justin Betteridge, Bryan Kisiel, Burr Settles, Estevam~R.
  Hruschka, and Tom~M. Mitchell. 2010.
\newblock Toward an architecture for never-ending language learning.
\newblock In \emph{Proceedings of the Twenty-Fourth AAAI Conference on
  Artificial Intelligence}, AAAI'10, page 1306–1313. AAAI Press.

\bibitem[{Chen et~al.(2019)Chen, Zhang, Zhang, Chen, and Chen}]{MetaR}
Mingyang Chen, Wen Zhang, Wei Zhang, Qiang Chen, and Huajun Chen. 2019.
\newblock \href {https://doi.org/10.18653/v1/D19-1431} {Meta relational
  learning for few-shot link prediction in knowledge graphs}.
\newblock In \emph{Proceedings of the 2019 Conference on Empirical Methods in
  Natural Language Processing and the 9th International Joint Conference on
  Natural Language Processing, {EMNLP-IJCNLP} 2019, Hong Kong, China, November
  3-7, 2019}, pages 4216--4225. Association for Computational Linguistics.

\bibitem[{Chopra et~al.(2005)Chopra, Hadsell, and LeCun}]{1467314}
S.~Chopra, R.~Hadsell, and Y.~LeCun. 2005.
\newblock \href {https://doi.org/10.1109/CVPR.2005.202} {Learning a similarity
  metric discriminatively, with application to face verification}.
\newblock In \emph{2005 IEEE Computer Society Conference on Computer Vision and
  Pattern Recognition (CVPR'05)}, volume~1, pages 539--546 vol. 1.

\bibitem[{Cormen et~al.(2009)Cormen, Leiserson, Rivest, and
  Stein}]{cormen2009introduction}
Thomas~H Cormen, Charles~E Leiserson, Ronald~L Rivest, and Clifford Stein.
  2009.
\newblock \emph{Introduction to algorithms}.
\newblock MIT press.

\bibitem[{Dasgupta et~al.(2018)Dasgupta, Ray, and Talukdar}]{HyTE}
Shib~Sankar Dasgupta, Swayambhu~Nath Ray, and Partha~P. Talukdar. 2018.
\newblock \href {https://doi.org/10.18653/v1/d18-1225} {Hyte: Hyperplane-based
  temporally aware knowledge graph embedding}.
\newblock In \emph{Proceedings of the 2018 Conference on Empirical Methods in
  Natural Language Processing, Brussels, Belgium, October 31 - November 4,
  2018}, pages 2001--2011. Association for Computational Linguistics.

\bibitem[{Dettmers et~al.(2018)Dettmers, Minervini, Stenetorp, and
  Riedel}]{ConvE}
Tim Dettmers, Pasquale Minervini, Pontus Stenetorp, and Sebastian Riedel. 2018.
\newblock \href
  {https://www.aaai.org/ocs/index.php/AAAI/AAAI18/paper/view/17366}
  {Convolutional 2d knowledge graph embeddings}.
\newblock In \emph{Proceedings of the Thirty-Second {AAAI} Conference on
  Artificial Intelligence, (AAAI-18), the 30th innovative Applications of
  Artificial Intelligence (IAAI-18), and the 8th {AAAI} Symposium on
  Educational Advances in Artificial Intelligence (EAAI-18), New Orleans,
  Louisiana, USA, February 2-7, 2018}, pages 1811--1818. {AAAI} Press.

\bibitem[{Devlin et~al.(2019)Devlin, Chang, Lee, and Toutanova}]{BERT}
Jacob Devlin, Ming{-}Wei Chang, Kenton Lee, and Kristina Toutanova. 2019.
\newblock \href {https://doi.org/10.18653/v1/n19-1423} {{BERT:} pre-training of
  deep bidirectional transformers for language understanding}.
\newblock In \emph{Proceedings of the 2019 Conference of the North American
  Chapter of the Association for Computational Linguistics: Human Language
  Technologies, {NAACL-HLT} 2019, Minneapolis, MN, USA, June 2-7, 2019, Volume
  1 (Long and Short Papers)}, pages 4171--4186. Association for Computational
  Linguistics.

\bibitem[{Garc{\'{\i}}a{-}Dur{\'{a}}n et~al.(2018)Garc{\'{\i}}a{-}Dur{\'{a}}n,
  Dumancic, and Niepert}]{TA-TransE}
Alberto Garc{\'{\i}}a{-}Dur{\'{a}}n, Sebastijan Dumancic, and Mathias Niepert.
  2018.
\newblock \href {https://aclanthology.org/D18-1516/} {Learning sequence
  encoders for temporal knowledge graph completion}.
\newblock In \emph{Proceedings of the 2018 Conference on Empirical Methods in
  Natural Language Processing, Brussels, Belgium, October 31 - November 4,
  2018}, pages 4816--4821. Association for Computational Linguistics.

\bibitem[{Goel et~al.(2020)Goel, Kazemi, Brubaker, and Poupart}]{DE-SimplE}
Rishab Goel, Seyed~Mehran Kazemi, Marcus Brubaker, and Pascal Poupart. 2020.
\newblock \href {https://aaai.org/ojs/index.php/AAAI/article/view/5815}
  {Diachronic embedding for temporal knowledge graph completion}.
\newblock In \emph{The Thirty-Fourth {AAAI} Conference on Artificial
  Intelligence, {AAAI} 2020, The Thirty-Second Innovative Applications of
  Artificial Intelligence Conference, {IAAI} 2020, The Tenth {AAAI} Symposium
  on Educational Advances in Artificial Intelligence, {EAAI} 2020, New York,
  NY, USA, February 7-12, 2020}, pages 3988--3995. {AAAI} Press.

\bibitem[{Han et~al.(2021)Han, Zhang, Ma, and Tresp}]{TKGCframework}
Zhen Han, Gengyuan Zhang, Yunpu Ma, and Volker Tresp. 2021.
\newblock \href {https://doi.org/10.18653/v1/2021.emnlp-main.639}
  {Time-dependent entity embedding is not all you need: {A} re-evaluation of
  temporal knowledge graph completion models under a unified framework}.
\newblock In \emph{Proceedings of the 2021 Conference on Empirical Methods in
  Natural Language Processing, {EMNLP} 2021, Virtual Event / Punta Cana,
  Dominican Republic, 7-11 November, 2021}, pages 8104--8118. Association for
  Computational Linguistics.

\bibitem[{Ji et~al.(2022)Ji, Pan, Cambria, Marttinen, and Yu}]{survey}
Shaoxiong Ji, Shirui Pan, Erik Cambria, Pekka Marttinen, and Philip~S. Yu.
  2022.
\newblock \href {https://doi.org/10.1109/TNNLS.2021.3070843} {A survey on
  knowledge graphs: Representation, acquisition, and applications}.
\newblock \emph{{IEEE} Trans. Neural Networks Learn. Syst.}, 33(2):494--514.

\bibitem[{Keet(2013)}]{Keet2013}
C.~Maria Keet. 2013.
\newblock \href {https://doi.org/10.1007/978-1-4419-9863-7_731} {\emph{Closed
  World Assumption}}, pages 415--415. Springer New York, New York, NY.

\bibitem[{Kim et~al.(2020)Kim, Hong, Ko, and Seo}]{MTL-KGC}
Bosung Kim, Taesuk Hong, Youngjoong Ko, and Jungyun Seo. 2020.
\newblock \href {https://doi.org/10.18653/v1/2020.coling-main.153} {Multi-task
  learning for knowledge graph completion with pre-trained language models}.
\newblock In \emph{Proceedings of the 28th International Conference on
  Computational Linguistics, {COLING} 2020, Barcelona, Spain (Online), December
  8-13, 2020}, pages 1737--1743. International Committee on Computational
  Linguistics.

\bibitem[{Kingma and Ba(2015)}]{DBLP:journals/corr/KingmaB14}
Diederik~P. Kingma and Jimmy Ba. 2015.
\newblock \href {http://arxiv.org/abs/1412.6980} {Adam: {A} method for
  stochastic optimization}.
\newblock In \emph{3rd International Conference on Learning Representations,
  {ICLR} 2015, San Diego, CA, USA, May 7-9, 2015, Conference Track
  Proceedings}.

\bibitem[{Lacroix et~al.(2020)Lacroix, Obozinski, and Usunier}]{TNTComplEx}
Timoth{\'{e}}e Lacroix, Guillaume Obozinski, and Nicolas Usunier. 2020.
\newblock \href {https://openreview.net/forum?id=rke2P1BFwS} {Tensor
  decompositions for temporal knowledge base completion}.
\newblock In \emph{8th International Conference on Learning Representations,
  {ICLR} 2020, Addis Ababa, Ethiopia, April 26-30, 2020}. OpenReview.net.

\bibitem[{Leblay and Chekol(2018)}]{TTransE}
Julien Leblay and Melisachew~Wudage Chekol. 2018.
\newblock \href {https://doi.org/10.1145/3184558.3191639} {Deriving validity
  time in knowledge graph}.
\newblock In \emph{Companion of the The Web Conference 2018 on The Web
  Conference 2018, {WWW} 2018, Lyon , France, April 23-27, 2018}, pages
  1771--1776. {ACM}.

\bibitem[{Lester et~al.(2021)Lester, Al-Rfou, and
  Constant}]{lester-etal-2021-power}
Brian Lester, Rami Al-Rfou, and Noah Constant. 2021.
\newblock \href {https://doi.org/10.18653/v1/2021.emnlp-main.243} {The power of
  scale for parameter-efficient prompt tuning}.
\newblock In \emph{Proceedings of the 2021 Conference on Empirical Methods in
  Natural Language Processing}, pages 3045--3059, Online and Punta Cana,
  Dominican Republic. Association for Computational Linguistics.

\bibitem[{Lewis et~al.(2020)Lewis, Liu, Goyal, Ghazvininejad, Mohamed, Levy,
  Stoyanov, and Zettlemoyer}]{BART}
Mike Lewis, Yinhan Liu, Naman Goyal, Marjan Ghazvininejad, Abdelrahman Mohamed,
  Omer Levy, Veselin Stoyanov, and Luke Zettlemoyer. 2020.
\newblock \href {https://doi.org/10.18653/v1/2020.acl-main.703} {{BART:}
  denoising sequence-to-sequence pre-training for natural language generation,
  translation, and comprehension}.
\newblock In \emph{Proceedings of the 58th Annual Meeting of the Association
  for Computational Linguistics, {ACL} 2020, Online, July 5-10, 2020}, pages
  7871--7880. Association for Computational Linguistics.

\bibitem[{Li et~al.(2018{\natexlab{a}})Li, Yang, Wang, Wang, Cui, and
  Zhang}]{li2018adaptive}
Bing Li, Xiaochun Yang, Bin Wang, Wei Wang, Wei Cui, and Xianchao Zhang.
  2018{\natexlab{a}}.
\newblock An adaptive hierarchical compositional model for phrase embedding.
\newblock In \emph{IJCAI}, pages 4144--4151.

\bibitem[{Li et~al.(2018{\natexlab{b}})Li, Ebrahimi~Kahou, Schulz, Michalski,
  Charlin, and Pal}]{NEURIPS2018_800de15c}
Raymond Li, Samira Ebrahimi~Kahou, Hannes Schulz, Vincent Michalski, Laurent
  Charlin, and Chris Pal. 2018{\natexlab{b}}.
\newblock \href
  {https://proceedings.neurips.cc/paper/2018/file/800de15c79c8d840f4e78d3af937d4d4-Paper.pdf}
  {Towards deep conversational recommendations}.
\newblock In \emph{Advances in Neural Information Processing Systems},
  volume~31. Curran Associates, Inc.

\bibitem[{Liu et~al.(2019)Liu, Ott, Goyal, Du, Joshi, Chen, Levy, Lewis,
  Zettlemoyer, and Stoyanov}]{RoBERTa}
Yinhan Liu, Myle Ott, Naman Goyal, Jingfei Du, Mandar Joshi, Danqi Chen, Omer
  Levy, Mike Lewis, Luke Zettlemoyer, and Veselin Stoyanov. 2019.
\newblock \href {http://arxiv.org/abs/1907.11692} {Roberta: {A} robustly
  optimized {BERT} pretraining approach}.
\newblock \emph{CoRR}, abs/1907.11692.

\bibitem[{Lovelace et~al.(2021)Lovelace, Newman-Griffis, Vashishth, Lehman, and
  Ros{\'e}}]{lovelace-etal-2021-robust}
Justin Lovelace, Denis Newman-Griffis, Shikhar Vashishth, Jill~Fain Lehman, and
  Carolyn Ros{\'e}. 2021.
\newblock \href {https://doi.org/10.18653/v1/2021.acl-long.82} {Robust
  knowledge graph completion with stacked convolutions and a student re-ranking
  network}.
\newblock In \emph{Proceedings of the 59th Annual Meeting of the Association
  for Computational Linguistics and the 11th International Joint Conference on
  Natural Language Processing (Volume 1: Long Papers)}, pages 1016--1029,
  Online. Association for Computational Linguistics.

\bibitem[{Lv et~al.(2022)Lv, Lin, Cao, Hou, Li, Liu, Li, and Zhou}]{fb15k-237n}
Xin Lv, Yankai Lin, Yixin Cao, Lei Hou, Juanzi Li, Zhiyuan Liu, Peng Li, and
  Jie Zhou. 2022.
\newblock \href {https://aclanthology.org/2022.findings-acl.282} {Do
  pre-trained models benefit knowledge graph completion? a reliable evaluation
  and a reasonable approach}.
\newblock In \emph{Findings of the Association for Computational Linguistics:
  ACL 2022}, pages 3570--3581, Dublin, Ireland. Association for Computational
  Linguistics.

\bibitem[{Niu et~al.(2021)Niu, Li, Tang, Geng, Dai, Liu, Wang, Sun, Huang, and
  Si}]{MTransH}
Guanglin Niu, Yang Li, Chengguang Tang, Ruiying Geng, Jian Dai, Qiao Liu, Hao
  Wang, Jian Sun, Fei Huang, and Luo Si. 2021.
\newblock \href {https://doi.org/10.1145/3404835.3462925} {Relational learning
  with gated and attentive neighbor aggregator for few-shot knowledge graph
  completion}.
\newblock In \emph{{SIGIR} '21: The 44th International {ACM} {SIGIR} Conference
  on Research and Development in Information Retrieval, Virtual Event, Canada,
  July 11-15, 2021}, pages 213--222. {ACM}.

\bibitem[{Paszke et~al.(2019)Paszke, Gross, Massa, Lerer, Bradbury, Chanan,
  Killeen, Lin, Gimelshein, Antiga, Desmaison, K{\"{o}}pf, Yang, DeVito,
  Raison, Tejani, Chilamkurthy, Steiner, Fang, Bai, and Chintala}]{pytorch}
Adam Paszke, Sam Gross, Francisco Massa, Adam Lerer, James Bradbury, Gregory
  Chanan, Trevor Killeen, Zeming Lin, Natalia Gimelshein, Luca Antiga, Alban
  Desmaison, Andreas K{\"{o}}pf, Edward~Z. Yang, Zachary DeVito, Martin Raison,
  Alykhan Tejani, Sasank Chilamkurthy, Benoit Steiner, Lu~Fang, Junjie Bai, and
  Soumith Chintala. 2019.
\newblock \href
  {https://proceedings.neurips.cc/paper/2019/hash/bdbca288fee7f92f2bfa9f7012727740-Abstract.html}
  {Pytorch: An imperative style, high-performance deep learning library}.
\newblock In \emph{Advances in Neural Information Processing Systems 32: Annual
  Conference on Neural Information Processing Systems 2019, NeurIPS 2019,
  December 8-14, 2019, Vancouver, BC, Canada}, pages 8024--8035.

\bibitem[{Petroni et~al.(2019)Petroni, Rockt{\"a}schel, Riedel, Lewis, Bakhtin,
  Wu, and Miller}]{petroni-etal-2019-language}
Fabio Petroni, Tim Rockt{\"a}schel, Sebastian Riedel, Patrick Lewis, Anton
  Bakhtin, Yuxiang Wu, and Alexander Miller. 2019.
\newblock \href {https://doi.org/10.18653/v1/D19-1250} {Language models as
  knowledge bases?}
\newblock In \emph{Proceedings of the 2019 Conference on Empirical Methods in
  Natural Language Processing and the 9th International Joint Conference on
  Natural Language Processing (EMNLP-IJCNLP)}, pages 2463--2473, Hong Kong,
  China. Association for Computational Linguistics.

\bibitem[{Raffel et~al.(2020)Raffel, Shazeer, Roberts, Lee, Narang, Matena,
  Zhou, Li, and Liu}]{JMLR:v21:20-074}
Colin Raffel, Noam Shazeer, Adam Roberts, Katherine Lee, Sharan Narang, Michael
  Matena, Yanqi Zhou, Wei Li, and Peter~J. Liu. 2020.
\newblock \href {http://jmlr.org/papers/v21/20-074.html} {Exploring the limits
  of transfer learning with a unified text-to-text transformer}.
\newblock \emph{Journal of Machine Learning Research}, 21(140):1--67.

\bibitem[{Saxena et~al.(2022)Saxena, Kochsiek, and Gemulla}]{KGT5}
Apoorv Saxena, Adrian Kochsiek, and Rainer Gemulla. 2022.
\newblock \href {https://doi.org/10.48550/arXiv.2203.10321}
  {Sequence-to-sequence knowledge graph completion and question answering}.
\newblock \emph{CoRR}, abs/2203.10321.

\bibitem[{Song et~al.(2019)Song, Tan, Qin, Lu, and Liu}]{MASS}
Kaitao Song, Xu~Tan, Tao Qin, Jianfeng Lu, and Tie{-}Yan Liu. 2019.
\newblock \href {http://proceedings.mlr.press/v97/song19d.html} {{MASS:} masked
  sequence to sequence pre-training for language generation}.
\newblock In \emph{Proceedings of the 36th International Conference on Machine
  Learning, {ICML} 2019, 9-15 June 2019, Long Beach, California, {USA}},
  volume~97 of \emph{Proceedings of Machine Learning Research}, pages
  5926--5936. {PMLR}.

\bibitem[{Sun et~al.(2019)Sun, Deng, Nie, and Tang}]{RotatE}
Zhiqing Sun, Zhi{-}Hong Deng, Jian{-}Yun Nie, and Jian Tang. 2019.
\newblock \href {https://openreview.net/forum?id=HkgEQnRqYQ} {Rotate: Knowledge
  graph embedding by relational rotation in complex space}.
\newblock In \emph{7th International Conference on Learning Representations,
  {ICLR} 2019, New Orleans, LA, USA, May 6-9, 2019}. OpenReview.net.

\bibitem[{Sun et~al.(2020)Sun, Vashishth, Sanyal, Talukdar, and
  Yang}]{reevaluation}
Zhiqing Sun, Shikhar Vashishth, Soumya Sanyal, Partha~P. Talukdar, and Yiming
  Yang. 2020.
\newblock \href {https://doi.org/10.18653/v1/2020.acl-main.489} {A
  re-evaluation of knowledge graph completion methods}.
\newblock In \emph{Proceedings of the 58th Annual Meeting of the Association
  for Computational Linguistics, {ACL} 2020, Online, July 5-10, 2020}, pages
  5516--5522. Association for Computational Linguistics.

\bibitem[{Toutanova and Chen(2015)}]{FB15k237}
Kristina Toutanova and Danqi Chen. 2015.
\newblock Observed versus latent features for knowledge base and text
  inference.
\newblock In \emph{Proceedings of the 3rd workshop on continuous vector space
  models and their compositionality}, pages 57--66.

\bibitem[{Trouillon et~al.(2016)Trouillon, Welbl, Riedel, Gaussier, and
  Bouchard}]{ComplEx}
Th{\'{e}}o Trouillon, Johannes Welbl, Sebastian Riedel, {\'{E}}ric Gaussier,
  and Guillaume Bouchard. 2016.
\newblock \href {http://proceedings.mlr.press/v48/trouillon16.html} {Complex
  embeddings for simple link prediction}.
\newblock In \emph{Proceedings of the 33nd International Conference on Machine
  Learning, {ICML} 2016, New York City, NY, USA, June 19-24, 2016}, volume~48
  of \emph{{JMLR} Workshop and Conference Proceedings}, pages 2071--2080.
  JMLR.org.

\bibitem[{Vashishth et~al.(2020)Vashishth, Sanyal, Nitin, and
  Talukdar}]{CompGCN}
Shikhar Vashishth, Soumya Sanyal, Vikram Nitin, and Partha~P. Talukdar. 2020.
\newblock \href {https://openreview.net/forum?id=BylA\_C4tPr}
  {Composition-based multi-relational graph convolutional networks}.
\newblock In \emph{8th International Conference on Learning Representations,
  {ICLR} 2020, Addis Ababa, Ethiopia, April 26-30, 2020}. OpenReview.net.

\bibitem[{Wang et~al.(2021{\natexlab{a}})Wang, Shen, Long, Zhou, Wang, and
  Chang}]{StAR}
Bo~Wang, Tao Shen, Guodong Long, Tianyi Zhou, Ying Wang, and Yi~Chang.
  2021{\natexlab{a}}.
\newblock \href {https://doi.org/10.1145/3442381.3450043} {Structure-augmented
  text representation learning for efficient knowledge graph completion}.
\newblock In \emph{{WWW} '21: The Web Conference 2021, Virtual Event /
  Ljubljana, Slovenia, April 19-23, 2021}, pages 1737--1748. {ACM} / {IW3C2}.

\bibitem[{Wang et~al.(2021{\natexlab{b}})Wang, Wood, Wan, Dras, and
  Johnson}]{wang-etal-2021-mention}
Yufei Wang, Ian Wood, Stephen Wan, Mark Dras, and Mark Johnson.
  2021{\natexlab{b}}.
\newblock \href {https://doi.org/10.18653/v1/2021.acl-long.9} {Mention flags
  ({MF}): Constraining transformer-based text generators}.
\newblock In \emph{Proceedings of the 59th Annual Meeting of the Association
  for Computational Linguistics and the 11th International Joint Conference on
  Natural Language Processing (Volume 1: Long Papers)}, pages 103--113, Online.
  Association for Computational Linguistics.

\bibitem[{Wang et~al.(2022)Wang, Xu, Sun, Hu, Tao, Geng, and
  Jiang}]{wang2022promda}
Yufei Wang, Can Xu, Qingfeng Sun, Huang Hu, Chongyang Tao, Xiubo Geng, and
  Daxin Jiang. 2022.
\newblock \href {https://doi.org/10.18653/v1/2022.acl-long.292} {{P}rom{DA}:
  Prompt-based data augmentation for low-resource {NLU} tasks}.
\newblock In \emph{Proceedings of the 60th Annual Meeting of the Association
  for Computational Linguistics (Volume 1: Long Papers)}, pages 4242--4255,
  Dublin, Ireland. Association for Computational Linguistics.

\bibitem[{Wang et~al.(2014)Wang, Zhang, Feng, and Chen}]{TransH}
Zhen Wang, Jianwen Zhang, Jianlin Feng, and Zheng Chen. 2014.
\newblock \href {http://www.aaai.org/ocs/index.php/AAAI/AAAI14/paper/view/8531}
  {Knowledge graph embedding by translating on hyperplanes}.
\newblock In \emph{Proceedings of the Twenty-Eighth {AAAI} Conference on
  Artificial Intelligence, July 27 -31, 2014, Qu{\'{e}}bec City, Qu{\'{e}}bec,
  Canada}, pages 1112--1119. {AAAI} Press.

\bibitem[{Wolf et~al.(2020)Wolf, Debut, Sanh, Chaumond, Delangue, Moi, Cistac,
  Rault, Louf, Funtowicz, Davison, Shleifer, von Platen, Ma, Jernite, Plu, Xu,
  Scao, Gugger, Drame, Lhoest, and Rush}]{huggingface}
Thomas Wolf, Lysandre Debut, Victor Sanh, Julien Chaumond, Clement Delangue,
  Anthony Moi, Pierric Cistac, Tim Rault, R{\'{e}}mi Louf, Morgan Funtowicz,
  Joe Davison, Sam Shleifer, Patrick von Platen, Clara Ma, Yacine Jernite,
  Julien Plu, Canwen Xu, Teven~Le Scao, Sylvain Gugger, Mariama Drame, Quentin
  Lhoest, and Alexander~M. Rush. 2020.
\newblock \href {https://doi.org/10.18653/v1/2020.emnlp-demos.6} {Transformers:
  State-of-the-art natural language processing}.
\newblock In \emph{Proceedings of the 2020 Conference on Empirical Methods in
  Natural Language Processing: System Demonstrations, {EMNLP} 2020 - Demos,
  Online, November 16-20, 2020}, pages 38--45. Association for Computational
  Linguistics.

\bibitem[{Xie et~al.(2022{\natexlab{a}})Xie, Wu, Shi, Zhong, Scholak, Yasunaga,
  Wu, Zhong, Yin, Wang, Zhong, Wang, Li, Boyle, Ni, Yao, Radev, Xiong, Kong,
  Zhang, Smith, Zettlemoyer, and Yu}]{xie2022unifiedskg}
Tianbao Xie, Chen~Henry Wu, Peng Shi, Ruiqi Zhong, Torsten Scholak, Michihiro
  Yasunaga, Chien{-}Sheng Wu, Ming Zhong, Pengcheng Yin, Sida~I. Wang, Victor
  Zhong, Bailin Wang, Chengzu Li, Connor Boyle, Ansong Ni, Ziyu Yao,
  Dragomir~R. Radev, Caiming Xiong, Lingpeng Kong, Rui Zhang, Noah~A. Smith,
  Luke Zettlemoyer, and Tao Yu. 2022{\natexlab{a}}.
\newblock \href {http://arxiv.org/abs/2201.05966} {Unifiedskg: Unifying and
  multi-tasking structured knowledge grounding with text-to-text language
  models}.
\newblock \emph{CoRR}, abs/2201.05966.

\bibitem[{Xie et~al.(2022{\natexlab{b}})Xie, Zhang, Li, Deng, Chen, Xiong,
  Chen, and Chen}]{GenKGC}
Xin Xie, Ningyu Zhang, Zhoubo Li, Shumin Deng, Hui Chen, Feiyu Xiong, Mosha
  Chen, and Huajun Chen. 2022{\natexlab{b}}.
\newblock \href {http://arxiv.org/abs/2202.02113} {From discrimination to
  generation: Knowledge graph completion with generative transformer}.
\newblock \emph{CoRR}, abs/2202.02113.

\bibitem[{Xiong et~al.(2018)Xiong, Yu, Chang, Guo, and Wang}]{GMatching}
Wenhan Xiong, Mo~Yu, Shiyu Chang, Xiaoxiao Guo, and William~Yang Wang. 2018.
\newblock \href {https://doi.org/10.18653/v1/d18-1223} {One-shot relational
  learning for knowledge graphs}.
\newblock In \emph{Proceedings of the 2018 Conference on Empirical Methods in
  Natural Language Processing, Brussels, Belgium, October 31 - November 4,
  2018}, pages 1980--1990. Association for Computational Linguistics.

\bibitem[{Xu et~al.(2019)Xu, Nayyeri, Alkhoury, Lehmann, and Yazdi}]{ATiSE}
Chengjin Xu, Mojtaba Nayyeri, Fouad Alkhoury, Jens Lehmann, and Hamed~Shariat
  Yazdi. 2019.
\newblock \href {http://arxiv.org/abs/1911.07893} {Temporal knowledge graph
  embedding model based on additive time series decomposition}.
\newblock \emph{CoRR}, abs/1911.07893.

\bibitem[{Xu et~al.(2020)Xu, Nayyeri, Alkhoury, Yazdi, and Lehmann}]{Tero}
Chengjin Xu, Mojtaba Nayyeri, Fouad Alkhoury, Hamed~Shariat Yazdi, and Jens
  Lehmann. 2020.
\newblock \href {https://doi.org/10.18653/v1/2020.coling-main.139} {Tero: {A}
  time-aware knowledge graph embedding via temporal rotation}.
\newblock In \emph{Proceedings of the 28th International Conference on
  Computational Linguistics, {COLING} 2020, Barcelona, Spain (Online), December
  8-13, 2020}, pages 1583--1593. International Committee on Computational
  Linguistics.

\bibitem[{Yang et~al.(2015)Yang, Yih, He, Gao, and Deng}]{DistMult}
Bishan Yang, Wen{-}tau Yih, Xiaodong He, Jianfeng Gao, and Li~Deng. 2015.
\newblock \href {http://arxiv.org/abs/1412.6575} {Embedding entities and
  relations for learning and inference in knowledge bases}.
\newblock In \emph{3rd International Conference on Learning Representations,
  {ICLR} 2015, San Diego, CA, USA, May 7-9, 2015, Conference Track
  Proceedings}.

\bibitem[{Yao et~al.(2019)Yao, Mao, and Luo}]{KG-BERT}
Liang Yao, Chengsheng Mao, and Yuan Luo. 2019.
\newblock \href {http://arxiv.org/abs/1909.03193} {{KG-BERT:} {BERT} for
  knowledge graph completion}.
\newblock \emph{CoRR}, abs/1909.03193.

\bibitem[{Zuo et~al.(2018)Zuo, Fang, Qian, Zhang, and Xu}]{DKRL}
Yukun Zuo, Quan Fang, Shengsheng Qian, Xiaorui Zhang, and Changsheng Xu. 2018.
\newblock \href {https://doi.org/10.1109/BigMM.2018.8499179} {Representation
  learning of knowledge graphs with entity attributes and multimedia
  descriptions}.
\newblock In \emph{Fourth {IEEE} International Conference on Multimedia Big
  Data, BigMM 2018, Xi'an, China, September 13-16, 2018}, pages 1--5. {IEEE}.

\end{thebibliography}
\bibliographystyle{acl_natbib}

\appendix

\section{Dataset}\label{appendix:dataset}

\paragraph{\dataicews} This dataset doesn't include any entity descriptions. As a result, we find the original data source\footnote{\url{https://dataverse.harvard.edu/dataverse/icews}} and create the description by combining ‘sector’ and ‘country’ entries for each entity.

\paragraph{\datanell} To conduct zero-shot learning for this dataset, we follow \citet{StAR} to reformat the raw dataset so that the relations in the dev/test sets do not appear in the train set. Additionally, we observe that textual representations of entities and relations are written in lower letters. To avoid pretrain-finetune data format mismatch, we further capitalize the surface words for each entity name. Dataset statistics are shown in Table \ref{tab:dataset}.
\begin{table}[!htbp]
	\centering
	\resizebox{\linewidth}{!}{
	\begin{tabular}{lcccccc}
\toprule
Dataset &Setting &$|\mathcal{E}|$ & $|\mathcal{R}|$ & |Train| & |Valid| &|Test|  \\
\midrule
\datawnrr &SKGC &40,943 &11 &86,835 &3,034 &3,134 \\
\datafb &SKGC &14,541 &237 &272,115 &17,535 &20,466 \\
\datafbn &SKGC &14,541 &93 &87,282 &7,041 &8,226 \\
\dataicews &TKGC &6,869 &230 &72,826 &8,941 &8,963 \\
\datanell &FKGC &68,544 &358 &189,635 &1,004 &2,158 \\
\bottomrule

\end{tabular}
}
\caption{Statistics of the Datasets.}
\label{tab:dataset} 
\end{table}

All of these datasets are open-source English-written sources without any offensive content. They are introduced only for research use. 


\section{Inplementation details}
\label{appendix:implementation}

We implement our \method using PyTorch~\cite{pytorch} and HuggingFace~\cite{huggingface}, and assess it on a single GPU (Tesla V100).

\paragraph{Model Input and Output} We follow the T5 standard unsupervised training paradigm. We form the query texts by masking the target entities with T5 default special tokens. Answer texts are also wrapped by the T5 special tokens. We use square brackets around descriptions to distinguish them from entity names. A special separation token ``|'' is inserted to separate entity, relation and meta-information. During the inference stage, our model generates the raw text, and we remove the wrapping special tokens and corresponding entity descriptions with regular expression, remaining the entity names as model predictions. Practical results suggest that the predicted entities can be determined by the entity names, so it is unnecessary to generate all the descriptions. Consequently, we perform an early stopping generation strategy, that is, the generation process will be stopped if the model outputs reach maximum entity name length.

\paragraph{Seq2Seq Dropout} Seq2Seq dropout is applied on the encoder input mask, randomly flipping the values from 1 to 0. Note that Seq2Seq dropout excludes the positions carrying special meanings, i.e. separation tokens, mask tokens and soft prompt.

\paragraph{Hyperparameters} In terms of hyperparameters, we select the batch size from \{32, 64, 128\}, learning rate from \{5e-3, 1e-3, 5e-4\}, description length from \{10, 40, 80\}, Seq2Seq dropout from \{0.0, 0.1, 0.2, 0.3\}. The optimal configurations are displayed in Table~\ref{tab:hyperparameters} 


\begin{table}[!htbp]
	\centering
	\resizebox{\linewidth}{!}{
	\begin{tabular}{lcccc}
\toprule
 &batch size &learning rate   &SRC/TGT desc. &S2S.Drop. \\
\midrule
\datawnrr &64 &1e-3 &40/40 &0.1 \\
\datafb &32 &1e-3 &80/80 &0.2 \\
\datafbn &32 &1e-3 &80/80 &0.2 \\
\dataicews &32 &5e-4 &40/40 &0.1 \\
\datanell &128 &5e-4 &0/0 &0.0 \\

\bottomrule
\end{tabular}
}
\caption{Hyperparameters for \method. SRC/TGT desc. denotes source and target description length. S2S.Drop denotes Seq2Seq dropout. }
\label{tab:hyperparameters} 
\end{table}

\end{document}